%% file: main.tex
\documentclass{article}
\usepackage{amssymb}

\usepackage[preprint]{corl_2025} 
\usepackage{graphicx}
\usepackage{amsmath}
\usepackage{amssymb} 
\usepackage{amsthm}
\usepackage{booktabs}
\usepackage{algorithm}
\usepackage{algorithmic}

\usepackage{booktabs} 
\usepackage{tabularx} 
\usepackage{xcolor}
\usepackage{threeparttable} 
\usepackage{pifont}
\usepackage[normalem]{ulem}  

\newcommand{\reffig}[1]{Figure \ref{#1}}
\newcommand{\refalg}[1]{Algorithm \ref{#1}}
\newcommand{\reftab}[1]{Table \ref{#1}}
\newcommand{\refeq}[1]{Equation (\ref{#1})}
\newcommand{\refeqn}[1]{(\ref{#1})}
\newcommand{\refsec}[1]{Section \ref{#1}}
\newcommand{\refapp}[1]{Appendix \ref{#1}}

\newcommand{\li}[1]{\textcolor{black}{#1}}
\newcommand{\zou}[1]{\textcolor{black}{#1}}

\title{ABPT: Amended Backpropagation through Time with Partially Differentiable Rewards}

\author{
  Fanxing Li\\
  SEIEE\\
  Shanghai Jiao Tong University, 
  China\\
  \texttt{li.fanxing@sjtu.edu.cn} \\
  \And
  Fangyu Sun \\
  SEIEE \\
  Shanghai Jiao Tong University \\
  \texttt{sunfly\_cc@sjtu.edu.cn} \\
  \AND
  Tianbao Zhang \\
  SEIEE \\
  Shanghai Jiao Tong University \\
  \texttt{zhangtianbao@sjtu.edu.cn} \\
  \And
  Danping Zou \\
  SEIEE \\
  Shanghai Jiao Tong University \\
  \texttt{dpzou@sjtu.edu.cn} \\
}

\begin{document}
\maketitle


\begin{abstract}
Quadrotor control policies can be trained with high performance using the exact gradients of the rewards to directly optimize policy parameters via backpropagation-through-time (BPTT). However, designing a fully differentiable reward architecture is often challenging. Partially differentiable rewards will result in biased gradient propagation that degrades training performance. To overcome this limitation, we propose Amended Backpropagation-through-Time (ABPT), a novel approach that mitigates gradient bias while preserving the training efficiency of BPTT. ABPT combines 0-step and N-step returns, effectively reducing the bias by leveraging value gradients from the learned Q-value function. Additionally, it adopts entropy regularization and state initialization mechanisms to encourage exploration during training. We evaluate ABPT on four representative quadrotor flight tasks \li{in both real world and simulation}. Experimental results demonstrate that ABPT converges significantly faster and achieves higher ultimate rewards than existing learning algorithms, particularly in tasks involving partially differentiable rewards. The code will be released at \url{Anonymous}.
\end{abstract}



\section{Introduction}
\input{introduction2}

\section{Related Work}
\input{relatedWork}

\input{methodology}

\section{Experiments}
\input{experiments}



\section{Conclusion}
\input{conclusion}


\clearpage
\appendix

\section{Proof}
\input{proof}

\section{Limitation}
\input{limitation}

\section{Task Defination}
\input{TaskDefination}

\section{Additional Experiment}
\subsection{Ablation Study}
\input{ablation}

\subsection{Reward Robustness}
\input{reward_robustness}

\subsection{Learning Rate Robustness}
\input{learning_rate_robustness}

\section{Discussion}
\input{discussion}
\clearpage
\section{Training Hyperparameters}
\input{training_paras}

\acknowledgments{If a paper is accepted, the final camera-ready version will (and probably should) include acknowledgments. All acknowledgments go at the end of the paper, including thanks to reviewers who gave useful comments, to colleagues who contributed to the ideas, and to funding agencies and corporate sponsors that provided financial support.}

\clearpage
\bibliography{reference}  

\end{document}

%% file: introduction2.tex
\zou{Quadrotors have demonstrated significant potential in various real-world applications including wild rescue, dangerous high-altitude work, and delivery. 
Recent work \cite{loquercio_learning_2021,loquercio_deep_2019, kaufmann_deep_2018} has shown end-to-end policies can be learned through imitation learning for controlling quadrotors from raw sensory data. However, the performance is largely restricted by expert's capability. Though reinforcement learning (RL) can tackle such limitation by self-exploration of agents, it relies on RL relies on zero-order gradient (ZOG) approximations \cite{sutton_reinforcement_2018} which require extensive sampling or replay mechanisms, which often results in slow convergence and sub-optimal training outcomes.}
Compared to imitation learning and RL algorithms, recent studies  \cite{zhang2024back,n_wiedemann_training_2023,lv2023sam,song2024learning, hu_seeing_2025} demonstrate that directly using first-order gradients for policy learning achieves faster convergence and superior performance, particularly for quadrotors \cite{zhang2024back,n_wiedemann_training_2023}. This is largely due to the good differentiable property of the quadrotor dynamics.

However, using first-order gradients for training requires not only the dynamics but also the reward function to be differentiable. While this condition is manageable for simple tasks, it becomes highly challenging for complex task objectives. Reward functions in such scenarios often involve non-differentiable components, such as conditional constants or binary rewards (e.g., awarding points upon task success), which violate differentiability requirements.
These non-differentiable elements disrupt the computation graph during backpropagation-through-time (BPTT), leading to biased first-order gradients. Notably, while first-order gradients remain unbiased under purely non-smooth dynamics \cite{zhang_adaptive_2023,suh_differentiable_2022}, they become biased when using non-differentiable rewards—a phenomenon we term \textbf{Biased Gradient}. This bias misguides training, causing optimization to stall in local minima and deviate from the intended direction of improvement.

\begin{figure}[h]
    \label{fig:real_world_0}
    \centering
    \includegraphics[width=1\textwidth]{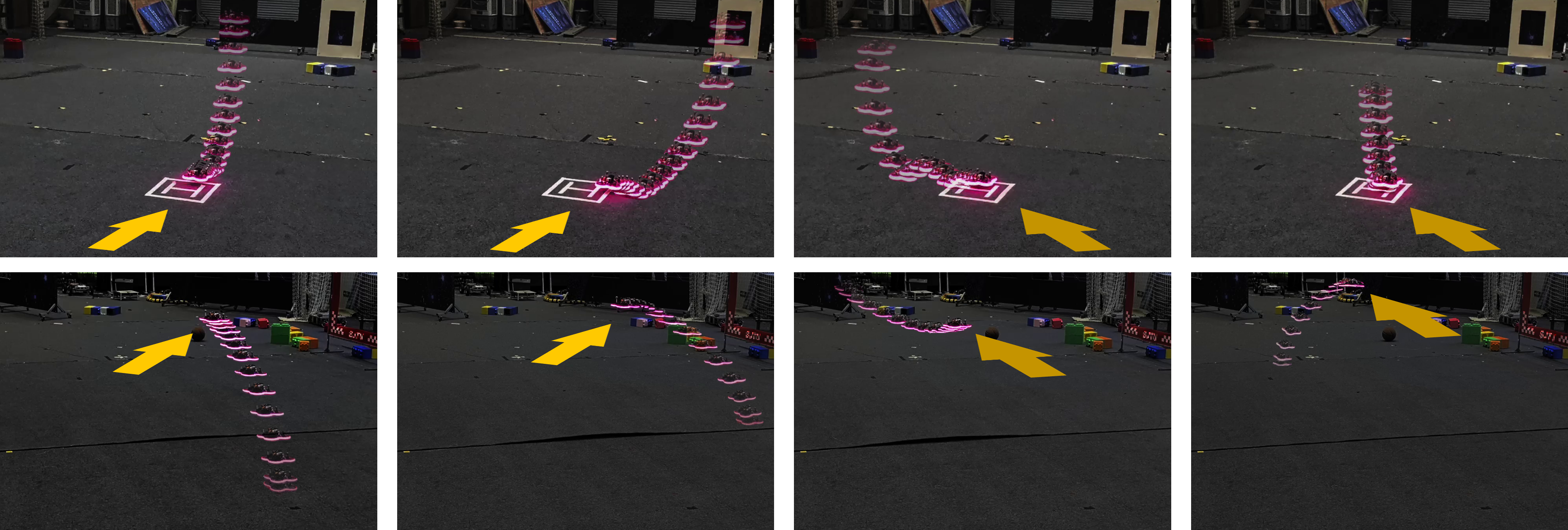}
    \caption{Our trained policies were validated in real world. 
    Two rows are separately four trials of landing (Top) and hovering (Bottom). 
    More videos are included in supplementary materials.
    }
\end{figure}

We propose an on-policy actor-critic approach - \textbf{Amended Backpropagation-through-Time (ABPT)} - 
to mitigate the bias gradient introduced by the non-differentiable rewards while keeping high policy learning performance in terms of training speed and converged rewards. Our approach combines 0-step returns with N-step returns \cite{sutton_reinforcement_2018}, leveraging value gradients generated by the 0-step returns to balance first-order gradient accuracy and exploitation. Additionally, ABPT incorporates entropy to counteract the excessive aggression of first-order gradient. It also employs a replay buffer to store state experiences, initializing episodes with these states to enhance sampling efficiency.
We evaluate our method on four representative quadrotor tasks, comparing it against classic policy gradient and first-order gradient methods. These tasks are designed to progressively increase the reward non-differentiability, testing the adaptability of each approach. Experimental results demonstrate that ABPT achieves the fastest convergence and highest final rewards across all baselines. This superiority is attributed to ABPT's ability to compensate for biased gradients and enhance exploration via entropy regularization and state replay. Furthermore, ABPT exhibits robustness across varying learning rates and reward structures.
Our technical contributions are summarized as follows:
\begin{itemize}
\item We propose ABPT, a novel approach to address the challenges in first-order gradient learning, including biased gradients caused by non-differentiable rewards and susceptibility to local minima.
\item We provide a comprehensive analysis of ABPT's effectiveness, offering insights to advance differentiable physics-based learning methods.
\item We introduce four representative quadrotor tasks as novel benchmark tests for evaluation. 
\zou{Our learned policies have been validated in real world as shown in \reffig{fig:real_world_0}}.
\end{itemize}

%% file: relatedWork.tex
\subsection{Reinforcement Learning}

Traditional reinforcement learning can be divided into two classes: model-free RL and model-based RL.
 Model-free RL includes value-based and policy-gradient methods. Value-based methods learn value functions to estimate long-term rewards. DQN \cite{mnih2013playing} introduced neural networks for discrete actions, while DDPG \cite{lillicrap2015continuous} extended this to continuous action spaces. TD3 \cite{fujimoto2018addressing} reduced overestimation bias with multiple value networks, and SAC \cite{haarnoja2018soft} used a maximum entropy framework for robust high-dimensional learning. Policy-gradient methods directly optimize policies using gradients. TRPO \cite{schulman2015trust} stabilized updates via trust regions, and PPO \cite{schulman2017proximal} simplified optimization with a clipped surrogate objective.

In contrast to model-free RL which treats the environment as a black box, model-based RL \cite{moerland_model-based_2023} introduces an additional process to learn the environment's dynamics. 
For example, PILCO~\cite{deisenroth2011pilco} and Dyna-Q~\cite{sutton1990integrated} leveraged  learned environment models to generate simulated experiences to accelerate training. Methods like \cite{chua2018deep,watter2015embed} employ trajectory sampling to plan over learned environment models. Dreamer \cite{hafner_dream_2019} embedded entire functions into a latent space, enabling end-to-end policy updates via backpropgation through time (BPTT). Despite their advantages, existing RL methods do not explicitly utilize the dynamics of robotics that can be precisely described by physical laws.

\subsection{Differentiable Simulators}
Policy learning via differentiable physics is an approach that integrates the physical simulations with differentiable dynamics to enable policy learning directly by using gradient based optimization. Making the dynamics differentiable in the simulator is the key to this approach. 
DiffTaichi \cite{hu_difftaichi_2020} is a comprehensive differentiable physics engine that includes simulations of fluid, gas, rigid body movement, and more. 
In the field of robotics, Brax \cite{freeman_brax_2021} offers differentiable versions of common RL benchmarks, built on four physics engines, including JAX and Mujoco \cite{todorov_mujoco_2012}. Another line of research focuses on addressing challenges in contact-rich environments. For example, Heiden et al. \cite{heiden_neuralsim_2021} tackle the contact-rich discontinuity problem in quadruped robots by employing a neural network to approximate the residuals. Dojo \cite{howell_dojo_2023} enhances contact solvers and integrates various integrators to accelerate computations while maintaining fidelity. Other research efforts focus on improving decision-making capabilities rather than control policies. VisFly \cite{li_visfly_2024} introduces a versatile drone simulator with fast rendering, based on Habitat-Sim \cite{savva2019habitat}, providing a platform for high-level applications. To enhance the efficiency, many simulators leverage GPU-accelerated frameworks like JAX \cite{schoenholz_jax_2020} and PyTorch \cite{paszke2017automatic} for faster computations.

 \subsection{First-order Gradient Training}
 With the differentiable simulators, the policy canbe trained through BPTT by using the first-order gradients. Though first-order gradients enables faster and more accurate gradient computation, accelerating policy training via BPTT \cite{mozer2013focused}, they suffer from gradient explosion/vanishing or instability caused by non-smooth dynamics. Many attempts try to address these issues and strengthen robustness. PODS \cite{mora_pods_2021} leverages both first- and second-order gradients with respect to cumulative rewards. SHAC \cite{xu_accelerated_2022} employs an actor-critic framework, truncates the learning window to avoid vanishing/exploding gradients, and smooths the gradient updates. AOBG \cite{suh_differentiable_2022} combines ZOG (policy gradient) with FOG, using an adaptive ratio based on gradient variance in the minibatch to avoid the high variance typical of pure FOG in discontinuous dynamics. AGPO \cite{gao_adaptive_gradient_2024} replaces ZOG in mixture with critic predictions, as Q-values offer lower empirical variance during policy rollouts. While both AGPO and AOBG converge to asymptotic rewards in significantly fewer timesteps, the mixture ratio requires excessive computational resources, leading to longer wall-time. 

Among these algorithms, SHAC \cite{xu_accelerated_2022} achieves training efficiency comparable to the naive first-order approach (BPTT). However, SHAC primarily focuses on challenges related to dynamics and overlooks the gradient bias introduced by partially differentiable rewards, which are common in real-world quadrotor tasks. These tasks often involve differentiable dynamics but rely on discrete or sparse reward signals, such as waypoint reaching events or minimum-time completions. Our approach directly tackles this limitation by introducing a 0-step return to address gradient bias from non-differentiable rewards and incorporating entropy to encourage exploration in challenging scenarios. The results highlight our method's superior performance in quadrotor applications, achieving the fastest convergence and the highest final rewards among all the evaluated methods.


%% file: methodology.tex
\section{Preliminaries}
The goal of reinforcement learning is to find a stochastic policy $\pi$ parameterized by $\theta$ that maximizes the expected cumulative reward or the expected return over a trajectory $\tau$, namely
\begin{equation}
    \arg\max_\theta \, \left\{\mathcal{J}_\theta[\tau] \triangleq \mathbb{E}_{\pi_\theta} \left[ \sum_{t=0}^{\infty}\gamma^t R(s_t) \right]\right\},
\end{equation}
where $R$ represents the reward function.
The expected return starting from a given state $s$, also named as  a state value function, is usually approximated by some parameters $\phi$, 
\begin{equation}
    V_{\phi}(s) = \mathcal{J}_\theta [\tau|s_0 = s].
\end{equation}
The action-value function $Q_{\phi}(s, a)$ indicates the expected return after taking action $a$, starting from state $s$.
The state-value function $V_\phi(s)$ can be expressed by the expected $Q_{\phi}(s,a)$ over the action space:
\begin{equation}
\label{eq:action_to_state}
    V_{\phi}(s) = \mathbb{E}_{a\sim \pi_\theta} \left[ Q_{\phi}(s, a) \right].
\end{equation}
In a common actor-critic pipeline, both the actor ($\pi_\theta$) and the critic ($V_{\phi}(s)$ or $Q_{\phi}(s, a)$) are approximated by neural networks. The key problem is how to estimate the actor gradients to optimize the expected return. The methods could be divided into two following categories: 

\textbf{Policy Gradient.}
Policy gradient deploys stochastic policies to sample trajectories conditioned on the policy's action probabilities. By mathematical derivation, it estimates the expected returns via log-probability. Given a batch of experience, policy gradient is computed by:
\begin{equation}
    \nabla^{[0]}_\theta \mathcal{J}_\theta = \frac{1}{|\mathcal{B}|} \left[ \sum_{\tau\in \mathcal{B}}\sum_{t=0}^T \nabla_\theta \log \pi_\theta(a_t \mid s_t) A^{\pi_\theta}(s_t, a_t) \right],
\end{equation}
where $A^{\pi_\theta}(\cdot)$ represents the advantage derived from the value functions using current policy, $\mathcal{B}$ denotes the minibatch of sampled trajectories, $\tau$ represents a trajectory within the minibatch. 
Because this process does not involve differentiable simulation, it is also named {zeroth-order gradient (ZOG)}.

\textbf{Value Gradient.}
Value gradient focuses on gradient computation through the action-value function:
\begin{equation}
    \nabla^{[q]}_\theta \mathcal{J}_\theta = \frac{1}{|\mathcal{B}|} \left[ \sum_{i=1}^{|\mathcal{B}|} \nabla_\theta Q_{\phi}\big(s^{i}, \pi_\theta(s^{i})\big) \right]
\end{equation}
\cite{gao_adaptive_gradient_2024} named this gradient estimator as {Q gradient (QG)}. Compared to {ZOG}, the performance of value approximation is extremely crucial to actor training. This is because {QG} directly relies on the back-propagation of the action-value function, while {ZOG} firstly estimates advantages among values according to current policy that enables actor training more robust to the critic.

\section{First-Order Gradient Approach with Non-differentiable Rewards}
Given the state dynamics $T$ and reward function $R$ being differentiable, one can compute the extract gradients of the expected return for policy learning via backpropagation through time. This exact gradient estimate is called as {first-order gradient (FOG)}:
\begin{equation}
\nabla_\theta \mathcal{J}_\theta = \left( \sum_{k=0}^{N-1} \gamma^k \frac{\partial R(s_{t+k})}{\partial \theta} \right),
\end{equation}
where $N$ represents the horizon length and $i$ denotes the $i$-th trajectory within the minibatch. To consider infinite return whiling avoiding gradient explosion, an approximated N-step return \cite{sutton_reinforcement_2018} has been introduced in \cite{xu_accelerated_2022}:
\begin{equation}
\label{eq:first_order_n_return}
\small 
\nabla_\theta \mathcal{J}_\theta = \left( \sum_{k=0}^{N-1} \gamma^k \frac{\partial R(s_{t+k})}{\partial \theta} \right) + \gamma^N \nabla_\theta V_{\phi}(s_{t+N}).
\end{equation}
Here, $V_\phi$ is the state-value function reparameterized by $\phi$. As shown in \cite{xu_accelerated_2022}, using this approximated N-step return can introduce smooth landscape for optimization and mitigate the gradient explosion issues. However, it cannot address non-differentiable rewards as we will discuss later.

\begin{figure}[h]
    \centering
    \includegraphics[width=0.7\linewidth]{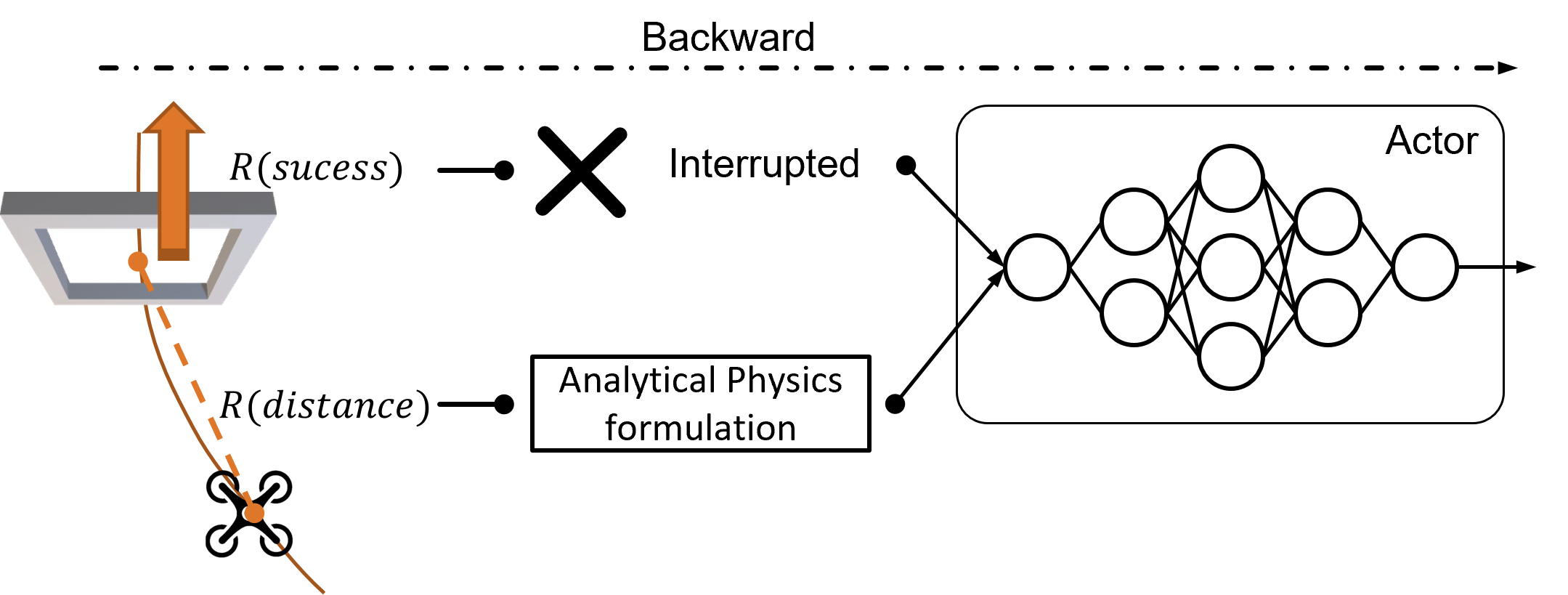}
    \caption{An illustration for explaining biased gradient. In a racing tasks for drone which we introduce in \refsec{sec:taskDefine}, the reward for passing the gate is a conditional constant, unable to automatically compute gradients.}
    \label{fig:biasGrad}
\end{figure}
\textbf{Biased Gradient.} 
When the rewards are partially differentiable, the gradients of non-differentiable part of the rewards will be absent from backpropgation.
For example, as shown in \reffig{fig:biasGrad}, a racing task's reward function consists of two components. The first one $R_{dist}$ depends on the distance from the drone to the gate to encourage the drone to move toward the gate, which is differentiable w.r.t the state. The second one $R_{succ}$ is a conditional constant reward given for successfully passing the gate, which does not involve gradient computation w.r.t. policy parameters. Therefore, although the desired objective involves both rewards
\begin{equation}
\small
\mathcal{J}_\theta = \left(\sum_{k=0}^{N-1} \gamma^t \Big(R_{dist}(s_{t+k})+R_{succ}(s_{t+k})\Big)\right)+\gamma^NV_\phi(s_{t+N}),
\end{equation}
backpropagation-through-time can effectively optimizes only the differentiable components.
As a result, the gate crossing reward $R_{succ}$, despite being crucial for learning the expected behavior (e.g. crossing the gate), is ignored during training. This ignorance can hinder the learned policy's ability to perform the desired actions. 


\section{The Proposed Method}
\begin{figure*}[h]
    \centering
    \includegraphics[width=1.0\textwidth]{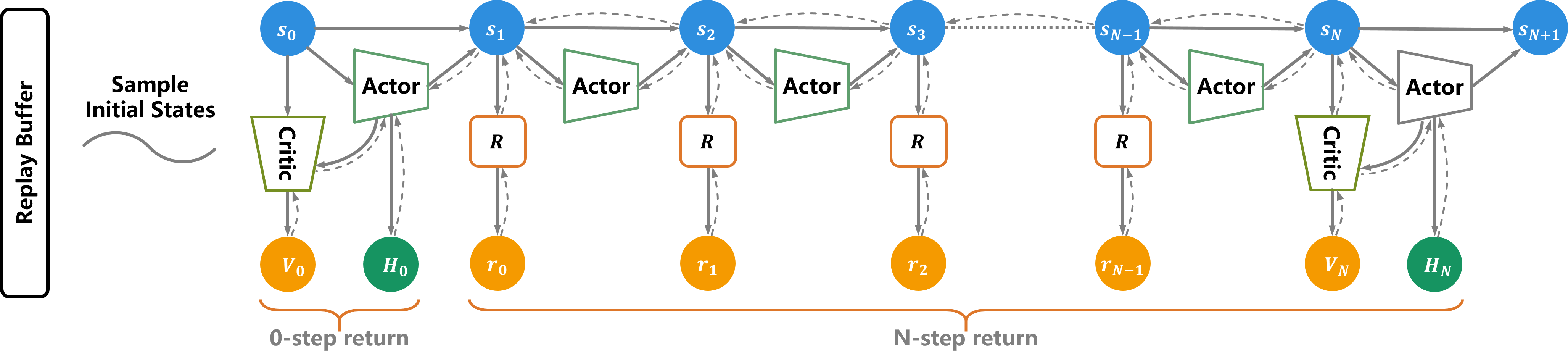}
    \caption{A overview of ABPT. ABPT combines 0-step return and n-step returns together, to compensate the biased gradient resulted by partially non-differentiable reward. The red dash lines indicate the direction of backpropagation. The replay buffer stores only visited states for episode initialization.}
    \label{fig:overview}
\end{figure*}

As previously discussed, explicit use of first-order gradients for policy learning requires addressing gradient bias caused by non-differentiable rewards. 
Motivated by the value gradient method, we propose to combine the 0-step return with $N$-step return for policy learning. 
Our method, \textbf{A}mended \textbf{B}ack\textbf{P}ropagation-through-\textbf{T}ime (ABPT), is an on-policy actor-critic learning approach. An overview is presented in \reffig{fig:overview}. 

During each training batch, we collect $|\mathcal{B}|$ trajectories with a horizon length $N$ and optimize the following objective function to update the actor network parameters $\theta$:

\begin{equation}
\label{eq:loss_actor}
\mathcal{J}_{\theta} = \frac{1}{2|\mathcal{B}|} \sum_{i=1}^{|\mathcal{B}|}  \Big(  \mathcal{J}_{\theta}^{N}+\mathcal{J}_{\theta}^{0} \Big)
\end{equation}
where $\mathcal{J}_{\theta}^{N}$, $\mathcal{J}_{\theta}^{0}$ are N-step return and 0-step return, defined as 
\begin{equation}
\small
\label{eq:n-return}
        \mathcal{J}_{\theta}^{N} = \underbrace{\left( \sum_{k = 0}^{N-1} \gamma^{k} {R}(s_{t+k}^i) \right) }_{\mathcal{G}^{t:t+N}_\theta}
        + (1-d)\gamma^N \underbrace{V_{\phi}(s_{t+N}^i)}_{\mathcal{V}_{\theta|\phi}^{t+N+1}},
        \:\:\:\:\; \; \; \; \; \; 
        \mathcal{J}_{\theta}^{0} =  \underbrace{V_{\phi}(s_{t}^i)}_{\mathcal{V}_{\theta|\phi}^{t}}
\end{equation}

Here, $d$ is a boolean variable indicating whether the current episode has ended, and $i$ denotes the trajectory index.  Given the trajectory is perform by $\pi_\theta$, each element is differentiable to parameter $\theta$. $\mathcal{G}_\theta^{t:t+N}$ represents the accumulated reward within the horizon and $\mathcal{V}_{\theta|\phi}^{t+N+1}$ is the value obtained by fixed critic. 
 Both 0-step return and N-step return are the expected value computed from the same action-value function $Q_\phi$. 
 \zou{We prove that using the objective function (\ref{eq:loss_actor}) for policy learning is equivalent to taking the average of the value gradient and the first-order gradient for backpropagation in \refapp{sec:proof}.}
 

\input{alg/alg1}
We use a Gaussian policy $\pi_\theta(a|s) = \mathcal{N}(\mu_\theta(s),\sigma_\theta(s))$ for the actor network and apply the reparameterization trick \cite{kingma2013auto} to gradient computation. We also normalize the actions using tanh function to stabilize the training process: 
\begin{equation}
\small
    a_t = \tanh(\mu_{\theta}(s_t) + \sigma_{\theta}(s_t) \epsilon), \quad \epsilon \sim \mathcal{N}(0, I).
\end{equation}
After updating the critic, target returns are estimated over time and used to train the value function by minimizing the MSE loss function:
\begin{equation}
\small
\label{eq:loss_critic}
\mathcal{L}_{\phi} = \mathbb{E}_{s \in \{\tau_i\}} \left\| V_{\phi}(s) - \tilde{V}_{\phi}(s) \right\|^2.
\end{equation}
We employ TD$-\lambda$ formulation \cite{sutton_reinforcement_2018} to estimate the expected return using exponentially averaging $k-$step returns:
\begin{equation}
\small
\label{eq:state_value_critic}
\tilde{V}_{\phi}(s_t) = (1 - \lambda) \left( \sum_{k=1}^{N - t - 1} \lambda^{k - 1} G_t^k \right) + \lambda^{N - t - 1} G_t^{N - t}
\end{equation}
where $G_t^k$ denotes $k-$step return from $t$:
\begin{equation}
\small
\label{eq:return}
G_t^k = \left( \sum_{l=0}^{k-1} \gamma^l r_{t+l} \right) + (1-d)\gamma^k V_{\phi}(s_{t+k}).
\end{equation}
The state-value function is derived from the action-value function :
\begin{equation}
\small
\label{eq:entropy}
V_{\phi}(s) = \mathbb{E}_{a \sim \pi} \left[ Q_{\phi}(s, a) \right] + \kappa H\left( \pi_\theta(\cdot \mid s) \right), 
\end{equation}
where we adopt an extra policy entropy term $H\left( \pi_\theta(\cdot \mid s) \right)$ to encourage exploration as in SAC \cite{haarnoja2018soft}. $\kappa$ is an adaptive ratio whose computation follows \cite{haarnoja2018soft}. To stabilize the critic training, we follow \cite{mnih2015human} to use a target critic $\phi^-$ to estimate the expected return (see \refeq{eq:state_value_critic}). 





 To further encourage exploration in policy learning, we additionally adopt a replay buffer to store all visited states during training. This buffer allows us to sample random, dynamically feasible states for episode initialization. While similar to the replay buffer in off-policy learning algorithms, our approach differs in that we store only visited states, not transitions, and use these states solely for initialization.
The pseudo code of the proposed method is shown in \refalg{alg:alg1}.

%% file: alg/alg1.tex
\begin{algorithm}[h]\
\caption{The proposed ABPT algorithm}
\label{alg:alg1}
\begin{algorithmic}[1]
\STATE Initialize parameters $\phi, \phi^-, \theta$ randomly.
\STATE Initialize state buffer $\mathcal{D}=\{\}$.

\WHILE{num time-steps $<$ total time-steps}
    \STATE \# \textit{Evaluate and collect states}
    \FOR {collecting steps = $1 \ldots i$}
        \STATE Add states $\mathcal{D} \leftarrow \mathcal{D} \cup \{(s_i)^N_{i=1}\}$ 
    \ENDFOR
    \STATE
    \STATE \# \textit{Train actor net}
    \STATE Sample minibatch $\{(s_i)\}_{\mathcal{B}} \sim \mathcal{D}$ as initial states
    \STATE Compute the gradient of $\mathcal{J}_{\theta}$ using \refeqn{eq:loss_actor}  and update the actor by one step of gradient ascent $\theta\leftarrow\theta+ \alpha \nabla_\theta \mathcal{J}_{\theta} \hfill $ 
    \STATE
    \STATE \# \textit{Train critic net}
    \STATE Compute the estimated value $\tilde{V}_\phi$ using \refeqn{eq:state_value_critic}
    \FOR {critic update step $c=1..C$}
        \STATE Compute the gradient of $\mathcal{L}_{\phi}$ using \refeqn{eq:loss_critic} and update weights by gradient descent $ \phi \leftarrow \phi - \alpha \nabla_\phi \mathcal{L}_{\phi}$ \hfill
        \STATE Softly update target critic $\phi^- \leftarrow (1 - \tau)\phi^- + \tau \phi$ \hfill 
    \ENDFOR

\ENDWHILE
\end{algorithmic}
\label{alg1}
\end{algorithm}

%% file: experiments.tex
We address the following questions in this section: 1) How does ABPT improve performance on typical quadrotor tasks compared to baseline methods? 2) What distinctive advantages does ABPT exhibit in behavior? 3) What is the contribution of each individual component? 

\subsection{Tasks for Evaluation}
\label{sec:taskDefine}
We conduct the evaluation on four quadrotor tasks, hovering, tracking, landing, racing, which involve different levels of complexity, where some of them are with non-differentiable rewards. Detailed task definitions are introduced in \refapp{sec:TaskDefination_app}. 


\subsection{Comparison with Baseline Methods}
We have compared our method with existing methods based on classic RL and differentiable physics. We present here the results of those on-policy methods, including PPO \cite{schulman2017proximal}, BPTT, and SHAC \cite{xu_accelerated_2022}.  As for off-policy methods, techniques such as SAC \cite{haarnoja2018softactorcriticoffpolicymaximum} struggle to demonstrate stable training curves, leading us to suspect that training the value network off-policy poses significant challenges for our quadrotor tasks. Additionally, we implemented mixed gradient methods \cite{suh_differentiable_2022,gao_adaptive_gradient_2024} without the original source code, but these algorithms require extremely long training times due to the need for variance estimation at each step.

\begin{figure*}[h]
    \centering
    \includegraphics[width=\textwidth]{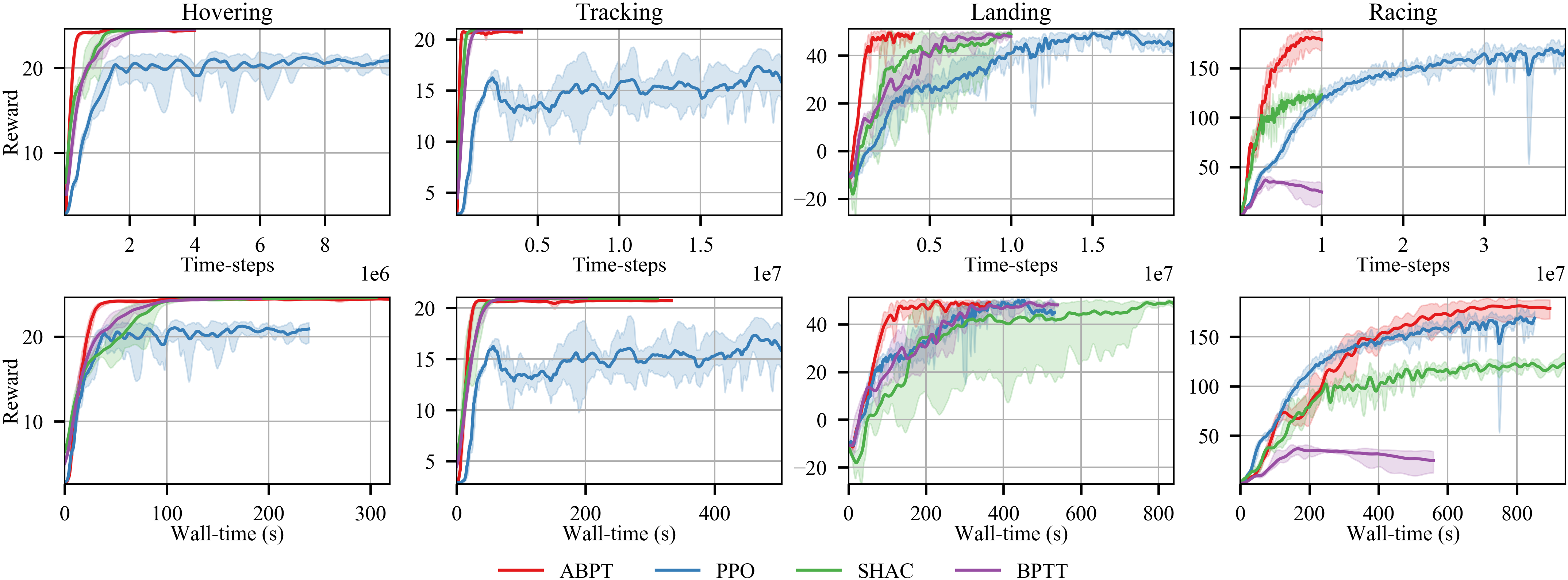}
    \caption{Training curves of PPO, SHAC, BPTT, and our ABPT in both time-step (\textbf{Top}) and wall-time (\textbf{Bottom}). Each curve is averaged over results from five random seeds, and the shaded area denotes the range of best and worst reward.}
    \label{fig:std_compa}
\end{figure*}

To ensure fair comparison, we implemented SHAC and BPTT by ourselves based on available source code, and adopt PPO from stable-baselines3 \cite{raffin2021stable} in VisFly simulator. All algorithms used parallel differentiable simulations to accelerate training. We tuned all hyperparameters to achieve optimal performance, and kept the settings consistent across all experiments as possible as we can. All experiments were conducted on the same laptop with an RTX 4090 GPU and a 32-core 13th Gen Intel(R) Core(TM) i9-13900K processor, with 5 random seeds for validation of robustness. Given the different time-step metrics across the algorithms, we compare their performance in terms of wall-time as well. Further details of the settings can be found in our source code (to be released). \reffig{fig:std_compa} provides reward curves of all methods during training.

\textbf{PPO}: 
PPO demonstrates moderate performance across the four tasks. However, due to the lack of an analytical gradient, PPO requires more sample collections to estimate the policy gradient, making it slower in terms of time-steps. In tasks that involve fully differentiable rewards such as hovering and tracking, it achieves the lowest asymptotic reward compared to FOG-based algorithms. As expected, PPO produces smooth and acceptable learning curves, since non-differentiable rewards do not impact the ZOG used by PPO.

\textbf{BPTT}: BPTT exhibits similar performance to SHAC and ABPT in the first two tasks. In the Landing task, despite the reward function incorporating non-differentiable discrete scores upon success, this component has only a minor impact on the FOG computation. This is because the reward function excluding this constant, has correctly determined the gradient via backpropagation. In the Racing task, we apply learning rate decay to BPTT, SHAC, and ABPT. BPTT shows the worst performance among all algorithms, demonstrating that the iteration quickly converges to a local minimum, caused by the bias introduced by the non-differentiable part in rewards. 

\textbf{SHAC}: Even though FOG is minimally biased in the Landing task, the curves from the five random seeds show significant fluctuations. The terminal success reward leads to high variance in the TD$-\lambda$ formulation used to estimate N-step returns, complicating critic training. As a result, SHAC performs worse than BPTT in the Landing task. In the Racing task, the terminal value partially addresses the non-differentiable components but still performs much worse than PPO and ABPT.

\textbf{Our ABPT}: In all tests, our ABPT method converges to the highest rewards. It achieves the fastest convergence speeds in the first three tasks and similar convergence speed to PPO in the racing tasks. 
By replaying visited states as initial states, ABPT enhances sampling efficiency by exploiting corner cases. 
Introducing the entropy helps suppress the high variance of the discrete reward space in the landing task, contributing to greater training stability. In the racing task, ABPT also outperforms PPO with a higher converged reward. This is largely due to that the value gradient introduced by 0-step returns is unaffected by non-differentiable rewards, making ABPT an effective method to compensate for biased gradient.

\subsection{Additional results}
\zou{We validate our trained policies in real-world experiments and provide a supplementary video demonstration. We further perform an ablation study on the key components of our approach (~\refapp{sec:ablation}) and evaluate its robustness to different kinds of rewards (~\refapp{sec:reward_robustness}) as well as different learning rates (~\refapp{sec:learning_rate_robustness}). 
}

%% file: conclusion.tex


We present a novel approach ABPT to training policies for quadrotor tasks robustly. It effectively addresses the challenges from the partially non-differentiable rewards associated with existing first-order gradient learning methods. We validated  ABPT on four quadrotor tasks — hovering, tracking, landing, and racing — and compared them with existing learning algorithms. The results show that ABPT achieves faster and more stable training processes and converges higher rewards across all tasks. ABPT is also robust to the learning rate and different kinds of rewards. The ablation study also shows the effectiveness of each key component of our approach. \zou{We also discuss the limitation of our approach in ~\refapp{sec:limitation} and present further discussion in ~\refapp{sec:discuss}.}

%% file: proof.tex
\label{sec:proof}
We explain how using the objective function (\ref{eq:loss_actor}) is equivalent to combining both the value gradient and the first-order gradient  for backpropagation. Supposed the value function $Q_\phi$ is well trained, the accumulated reward within the horizon can be approximated as: 
 \begin{equation}
 \small
     \mathcal{G}_\theta^{t:t+N} \approx\mathcal{V}_{\theta|\phi}^{t}-(1-d)\gamma^N \mathcal{V}_{\theta|\phi}^{t+N+1}.
 \end{equation}
 Its value gradient is then given by 
 \begin{equation}
 \small
     \nabla_\theta^{[q]} \mathcal{G}_\theta^{t:t+N} = \nabla_\theta\mathcal{V}_{\theta|\phi}^{t} - (1-d)\gamma^N  \nabla_\theta \mathcal{V}_{\theta|\phi}^{t+N+1}
 \end{equation}

regardless of the differentiability of the rewards. Noting that, unlike \cite{xu_accelerated_2022},  we specifically use action-value function $Q_\phi$ to compute the value to ensure $\mathcal{G}_\theta^{t:t+N}$ is differentiable with respect to $\theta$, which makes this derivative expression meaningful, otherwise the derivative would be zero if using $V_\phi$ solely with state input. Let $\nabla_\theta\mathcal{G}_\theta^{t:t+N}$ denote the first-order gradient of the accumulated reward. The average of the two gradients can be expressed as:
\begin{equation}
\small
\bar{\nabla}_\theta \mathcal{G}_\theta^{t:t+N} = \frac{1}{2}\left(\nabla^{[q]}_\theta \mathcal{G}_\theta^{t:t+N} + \nabla_\theta \mathcal{G}_\theta^{t:t+N}\right).
\end{equation}
It is straightforward to verify that taking the derivative of \refeqn{eq:loss_actor} yields the following gradient for backpropagation:

\begin{equation} 
\label{eq:combined_gradients}
\small
\begin{aligned}
& \nabla_\theta \mathcal{J}_\theta = \frac{1}{|\mathcal{B}|}\sum_{i=1}^{|\mathcal{B}|}\left[ \bar{\nabla}_\theta \mathcal{G}_\theta^{t:t+N+}+(1-d)\gamma^N \nabla_\theta \mathcal{V}_{\theta|\phi}^{t+N+1} \right] \\
& = \frac{1}{2|\mathcal{B}|}\sum_{i=1}^{|\mathcal{B}|} \left[\underbrace{\nabla^{[q]}_\theta \mathcal{G}_\theta^{t:t+N}}_{\nabla\mathcal{J}^{0}_\theta} + \underbrace{\nabla_\theta \mathcal{G}_\theta^{t:t+N}+(1-d)\gamma^N \nabla_\theta \mathcal{V}_{\theta|\phi}^{t+N+1}}_{\nabla\mathcal{J}^N_\theta} \right].
\end{aligned}
\end{equation}

Therefore, the difference between this gradient and the gradient (\ref{eq:first_order_n_return}) used in \cite{xu_accelerated_2022}  is that the first-order gradients in \refeqn{eq:first_order_n_return} are combined with the value gradients. By leveraging this combination, our method remains effective in guiding the parameter updates toward the correct direction, when the first-order gradient is biased due to the non-differentiable rewards.

\begin{figure}[h]
    \centering
    \includegraphics[width=0.5\linewidth]{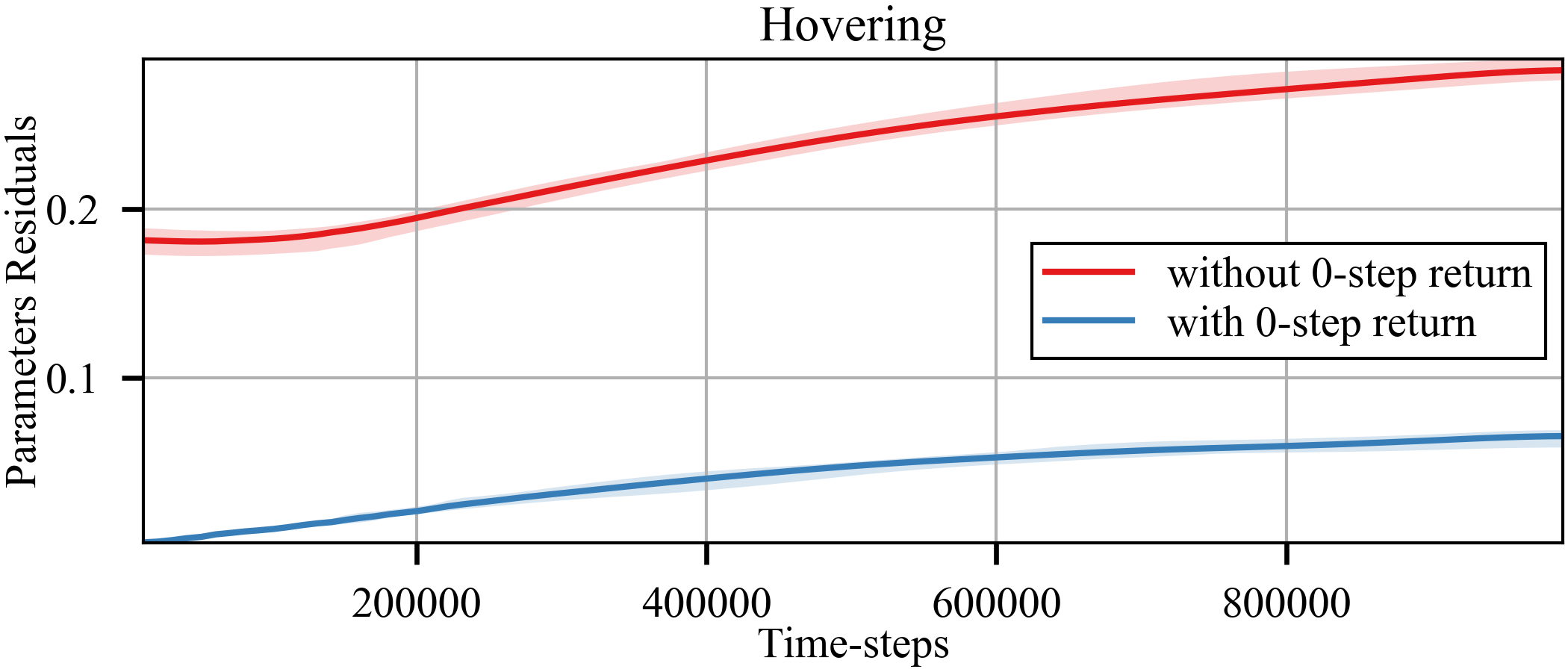}
    \caption{ The curve shows the the difference between the parameters trained with fully differentiable and partially differentiable rewards. We deliberately detach parts of the rewards to interrupt gradient backpropagation and retrain the policy with or without combining the 0-step return. }
    \label{fig:detach_diff}
\end{figure}

We conduct a simple experiment to assess the effectiveness of incorporating the 0-step return in addressing gradient bias. We deliberately detach parts of rewards in the hovering task (see \refsec{sec:taskDefine}) to mimic non-differentiable rewards,then back-propagate to compute gradient of network parameter. As shown in \reffig{fig:detach_diff}, combining the 0-step return with the N-step return in the objective function \refeqn{eq:loss_actor} for training significantly reduces the model parameter residuals.

%% file: limitation.tex
\label{sec:limitation}
\li{
ABPT effectively enhances the efficiency and robustness of training processes utilizing analytical gradients, even in scenarios involving partially differentiable reward structures.
However, while it significantly mitigates the gradient bias caused by non-differentiable reward components, it may still fail to fully eliminate extreme bias if the biased gradient is excessively large. In other words, if the non-differentiable rewards contribute too heavily to the overall reward, ABPT's training performance may also degrade. Therefore, we strongly recommend that, when designing reward functions, priority should be given to incorporating smooth and differentiable variables to the greatest extent possible. In the following work, we will further explore how to adaptively mix the gradient while avoiding cost too much resources on mixture ratio computation.
}

%% file: TaskDefination.tex
\label{sec:TaskDefination_app}

\textbf{Hovering.} Starting from a random position, the quadrotor needs to hover stably at a target location. Fully differentiable rewards are used in this task.

\textbf{Tracking.} Starting from a random position, the quadrotor tracks a circular trajectory with a fixed linear velocity. Fully differentiable rewards are used in this task.

\textbf{Landing.} Starting from a random position, the quadrotor gradually descends, and eventually lands at the required position on the ground. This task involves using non-differentiable rewards during training. 

\textbf{Racing.} The quadrotor flies through four static gates as quickly as possible in a given order repeatedly. This task involves more rewards with some of them non-differentiable. 

We use the quadrotor simulator VisFly \cite{li_visfly_2024} as our training environment, where the quadrotor dynamics are well implemented with automatic FOG computation achieved via \cite{paszke2017automatic}.A comprehensive description of observation and reward structure is presented in \reftab{tab:taskDefine}.
\begin{figure*}[ht]
    \centering
    \includegraphics[width=\linewidth]{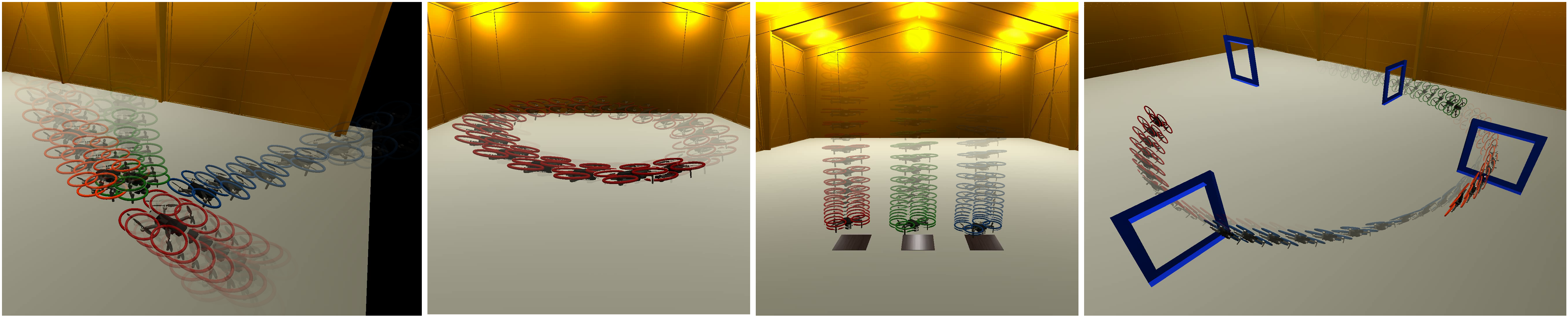}
    \caption{Quadrotor tasks (left to right): hovering, tracking, landing, and racing. We illustrate multiple drones (in different colors) simultaneously to indicate episodes from different initial states. }
    \label{fig:demo}
\end{figure*}
\input{tables/envSetting}

%% file: tables/envSetting.tex
\begin{table*}[h]
\scriptsize
\centering
\caption{Observations and rewards used in benchmark quadrotor tasks}
\label{tab:taskDefine}
\begin{tabularx}{\textwidth}{l  | X | l }
\toprule
\textbf{Environments} & \textbf{Observation} & \textbf{Reward Function}  \\
\midrule
\textbf{Hovering}  & state \& $\hat{p}$ & $c - k_1 \left \|p-\hat{p}\right\| - k_2 \left\|q-\hat{q}\right\| - k_3 \left\|v\right\| - k_4 \left\|\omega\right\|$ (fully DIFF) \\
\textbf{Tracking} & state \& next 10 $\hat{p}_{i=1\sim10}$  & $c- k_1 \left \|p-\hat{p}_0\right\| - k_2 \left\|q-\hat{q}\right\| - k_3 \left\|v\right\| - k_4 \left\|\omega\right\|$ (fully DIFF)\\
\textbf{Landing} & state \& $\hat{p}$ & $-k_1 f^+ \big( \left \|p_{xy}-\hat{p}_{xy}\right\| \big) + k_2 f^+ \big(  \left\|v_z-\hat{v}_z\right\| \big) + k_3 s$ (partially DIFF)\\
\textbf{Racing} & state \& next 2 $\hat{p}_{i=1,2}$ of gates & $c - k_1 \left \|p-\hat{p}_0\right\| - k_2 \left\|q-\hat{q}\right\| - k_3 \left\|v\right\| - k_4 \left\|\omega\right\| + k_5s$ (partially DIFF) \\
\bottomrule
\end{tabularx}

\begin{tablenotes}
\item[1] $c$ represents a small constant used to ensure the agent remains alive. $k_i$ denotes constant weights for different reward contributions, with these weights being distinctly defined for each task. $s$ is a boolean variable that indicates whether the task is successfully completed. The state comprises position ($p$), orientation ($q$), linear velocity ($v$), and angular velocity ($\omega$). $f^+(\cdot)$ and $\hat{(\cdot)}$ are separately an increasing mapping function used to normalize the reward and target status. DIFF is abbreviation for differentiable. All the action types are individual thrusts.

\end{tablenotes}

\end{table*}

%% file: ablation.tex
\begin{figure*}[ht]
    \centering
    \includegraphics[width=\textwidth]{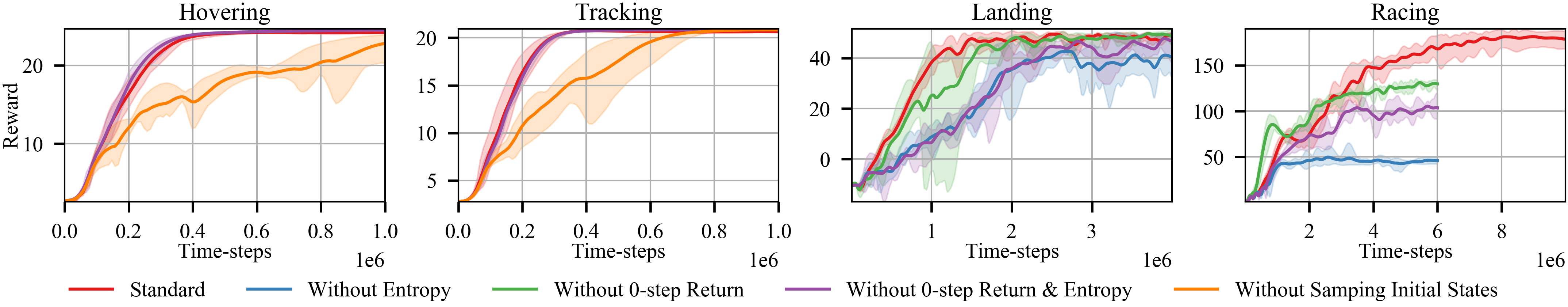}
    \caption{Ablation study: the key components of ABPT are sequentially removed in turn to evaluate each one's contribution.}
    \label{fig:ablation}
\end{figure*}

\label{sec:ablation}
We evaluate the effectiveness of key components of our approach by removing each during training. As shown in \reffig{fig:ablation}, the results show that : 
1) Incorporating 0-step return clearly improves the training performance in tasks with non-differentiable rewards such as landing and racing. 2) Adding entropy regularization (see \refeq{eq:entropy}) helps escape from the local minima, achieving higher converged rewards , particularly in landing and racing tasks. 3) Initializing episodes from previously visited states stored in the buffer enhances sampling efficiency, accelerating convergence, as observed in the first two tasks.


%% file: reward_robustness.tex
\label{sec:reward_robustness}

\begin{figure}[ht]
    \centering
    \includegraphics[width=0.5\linewidth]{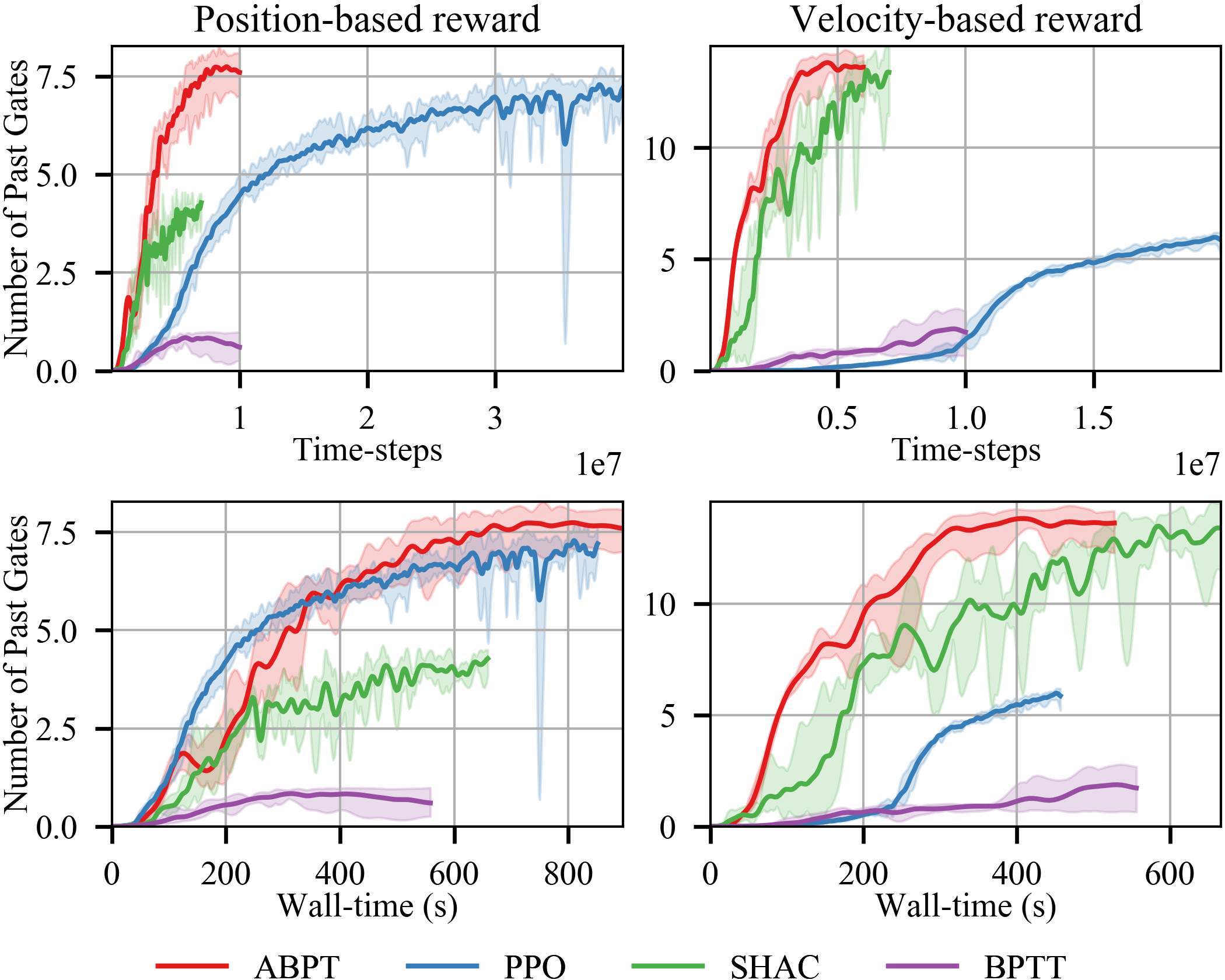}
    \caption{Training curves with different rewards: position-based rewards (\textbf{Left column}) and velocity-based rewards (\textbf{Right column}). The number of passed gates is visualized as the performance metric because of different rewards used for training.
    }
    \label{fig:rewardrobust}
\end{figure}

Designing an appropriate reward function is highly challenging for real-world applications, particularly when dealing with specific requirements. Ensuring robustness to reward architecture is crucial for the training algorithms. In the racing task, we redefined the reward function by replacing Euclidean distance with approaching velocity in the reward.
As shown in \reffig{fig:rewardrobust}, ABPT outperforms other methods with both position-based and velocity-based rewards. With fewer non-differentiable components, velocity-based rewards allow ABPT and SHAC to pass more gates per episode, while BPTT fails due to gradient issues.

%% file: learning_rate_robustness.tex
\label{sec:learning_rate_robustness}
\begin{figure*}[ht]
    \centering
    \includegraphics[width=\textwidth]{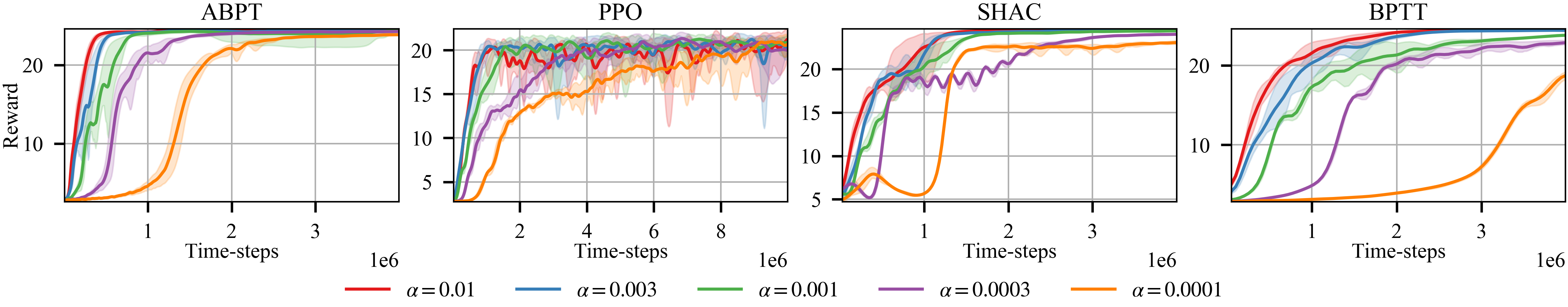}
    \caption{Training curves with different learning rates $0.01,0.003,0.001,0.0003,0.00001$. The proposed ABPT exhibit stable and fast training performance in all learning rates. 
    }
    \label{fig:precise}
\end{figure*}

We evaluated the training performance using different learning rates. The fully differentiable hovering task is used for evaluation.
As shown in \reffig{fig:precise}, the proposed ABPT exhibit stable and fast
training performance in all learning rates. PPO has the highest variance compared to other FOG-based algorithms, as expected demonstrating that FOG is much precise than ZOG. 
Increasing learning rate has slight improvement on acceleration once it surpass 0.001 for PPO and SHAC, while ABPT's convergence speed stably grows with increasing learning rate. 

%% file: discussion.tex
\label{sec:discuss}
The authors in \cite{gao_adaptive_gradient_2024}  address the limitations of first-order gradients, such as bias under discontinuities and high variance in stiff contacts, by adaptive mixing policy gradients and first-order gradients at every time step. While probably effective in reducing the gradient bias from non-differentiable rewards, this method incurs significant computational overhead, limiting its practicality. In contrast, we mitigate the gradient bias by simply averaging the value gradient and the first-order gradient. This approach is far more efficient, requiring minimal overhead for evaluating the value gradients only at the initial and the last time steps.

We also explored incorporating $k$-step value functions ($k=0,\ldots,N-1$) at each step within a finite horizon, following a similar approach to \cite{suh_differentiable_2022}. However, this led to significant fluctuations in the learning curves.

%% file: training_paras.tex
Tables 2$\sim$5 contain the parameters for the baseline experiments, Tables 6$\sim$8 for the ablation experiments, and Table 9 for the reward robustness experiments.
\input{tables/shac_paras}

\input{tables/ppo_paras}

\input{tables/bptt_paras}
\input{tables/mtd_paras}

\input{tables/mtd_ablation}

\input{tables/r_robust}

%% file: tables/shac_paras.tex
\begin{table*}[ht!]
\scriptsize
\centering
\caption{Hyerparameters of SHAC}
\label{tab:comparision}
\begin{tabular}{l |  l | l | l | l}
\toprule
& \textbf{Hovering} & \textbf{Tracking} & \textbf{Landing} & \textbf{Racing} \\
\midrule
learning rate $\alpha$ & 0.01 & 0.01 & 0.01 & 0.002  \\
number of parallel environments $n$ & 100 & 100 & 100 & 100  \\
discount factor $\gamma$ & 0.99 & 0.99 & 0.99 & 0.99  \\
training critic steps per minibatch & 10 & 10 & 10 & 10 \\ 
weight decay & 0.00001 & 0.00001  & 0.00001 & 0.00001 \\ 
target critic update factor $\tau$ & 0.005 & 0.005 & 0.005 & 0.005 \\ 
decay learning rate & False & False & False & True \\ 
value estimation factor $\lambda$  & 0.95 &  0.95 &  0.95 &  0.95 \\
horizon length $H$ & 96 & 96 & 96 & 96 \\
Optimizer & Adam & Adam & Adam & Adam  \\
\bottomrule
\end{tabular}

\end{table*}

%% file: tables/ppo_paras.tex
\begin{table*}[!h]
\scriptsize
\centering
\caption{Hyerparameters of PPO}
\label{tab:comparision}
\begin{tabular}{l |  l | l | l | l}
\toprule
& \textbf{Hovering} & \textbf{Tracking} & \textbf{Landing} & \textbf{Racing} \\
\midrule
learning rate $\alpha$ & 0.001 & 0.0002 & 0.0005 & 0.001  \\
number of parallel environments $n$ & 100 & 100 & 100 & 100  \\
discount factor $\gamma$ & 0.99 & 0.99 & 0.99 & 0.99  \\
minibatch size & 25600 & 25600 & 25600 & 51200  \\
training critic steps per minibatch & 5 & 5 & 5 & 5 \\ 
weight decay & 0.00001 & 0.00001  & 0.00001 & 0.00001 \\ 
GAE $\lambda$ & 1 & 1 & 1 & 1  \\
Optimizer & Adam & Adam & Adam & Adam  \\
\bottomrule
\end{tabular}
\end{table*}

%% file: tables/bptt_paras.tex
\begin{table*}[h!]
\scriptsize
\centering
\caption{Hyerparameters of BPTT}
\label{tab:comparision}
\begin{tabular}{l | l | l | l | l}
\toprule
& \textbf{Hovering} & \textbf{Tracking} & \textbf{Landing} & \textbf{Racing} \\
\midrule
learning rate $\alpha$ & 0.01 & 0.01 & 0.005 & 0.002  \\
number of parallel environments $n$ & 100 & 100 & 100 & 100  \\
discount factor $\gamma$ & 0.99 & 0.99 & 0.99 & 0.99  \\
weight decay & 0.00001 & 0.00001  & 0.00001 & 0.00001 \\ 
decay learning rate & False & False & False & True \\ 
horizon length $H$ & 256 & 256 & 256 & 512 \\
Optimizer & Adam & Adam & Adam & Adam  \\
\bottomrule
\end{tabular}
\end{table*}

%% file: tables/mtd_paras.tex
\begin{table*}[h!]
\scriptsize
\centering
\caption{Hyerparameters of ABPT}
\label{tab:comparision}
\begin{tabular}{l |  l | l | l | l}
\toprule
& \textbf{Hovering} & \textbf{Tracking} & \textbf{Landing} & \textbf{Racing} \\
\midrule
learning rate $\alpha$ & 0.01 & 0.01 & 0.01 & 0.01  \\
number of parallel environments $n$ & 100 & 100 & 100 & 100  \\
discount factor $\gamma$ & 0.99 & 0.99 & 0.99 & 0.99  \\
training critic steps per minibatch & 10 & 10 & 10 & 10 \\ 
weight decay & 0.00001 & 0.00001  & 0.00001 & 0.00001 \\ 
target critic update factor $\tau$ & 0.005 & 0.005 & 0.005 & 0.005 \\ 
decay learning rate & False & True & False & True \\ 
value estimation factor $\lambda$  & 0.95 &  0.95 &  0.95 &  0.95 \\
horizon length $H$ & 96 & 96 & 96 & 96 \\
replay buffer size & 1000000 & 1000000 & 1000000 & 50000 \\
Optimizer & Adam & Adam & Adam & Adam  \\

\bottomrule
\end{tabular}

\end{table*}

%% file: tables/mtd_ablation.tex
\begin{table*}[h!]
\scriptsize
\centering
\caption{Hyerparameters of ABPT no 0-step Value}
\label{tab:comparision}
\begin{tabular}{l |  l | l | l | l}
\toprule
& \textbf{Hovering} & \textbf{Tracking} & \textbf{Landing} & \textbf{Racing} \\
\midrule
learning rate $\alpha$ & 0.01 & 0.01 & 0.01 & 0.01  \\
number of parallel environments $n$ & 100 & 100 & 100 & 100  \\
discount factor $\gamma$ & 0.99 & 0.99 & 0.99 & 0.99  \\
training critic steps per minibatch & 10 & 10 & 10 & 10 \\ 
weight decay & 0.00001 & 0.00001  & 0.00001 & 0.00001 \\ 
target critic update factor $\tau$ & 0.005 & 0.005 & 0.005 & 0.005 \\ 
decay learning rate & False & True & False & True \\ 
value estimation factor $\lambda$  & 0.95 &  0.95 &  0.95 &  0.95 \\
horizon length $H$ & 96 & 96 & 96 & 96 \\
replay buffer size & 1000000 & 1000000 & 1000000 & 50000 \\
Optimizer & Adam & Adam & Adam & Adam  \\
\bottomrule
\end{tabular}
\end{table*}

\begin{table*}[ht!]
\scriptsize
\centering
\caption{Hyerparameters of ABPT no Entropy}
\label{tab:comparision}
\begin{tabular}{l |  l | l | l | l}
\toprule
& \textbf{Hovering} & \textbf{Tracking} & \textbf{Landing} & \textbf{Racing} \\
\midrule
learning rate $\alpha$ & 0.01 & 0.01 & 0.002 & 0.002  \\
number of parallel environments $n$ & 100 & 100 & 100 & 100  \\
discount factor $\gamma$ & 0.99 & 0.99 & 0.99 & 0.99  \\
training critic steps per minibatch & 10 & 10 & 10 & 10 \\ 
weight decay  & 0.00001 & 0.00001 & 0.00001 & 0.00001 \\ 
target critic update factor $\tau$ & 0.005 & 0.005 & 0.005 & 0.005 \\ 
decay learning rate  & False & True & False & True \\ 
value estimation factor $\lambda$ &  0.95 &  0.95 &  0.95 &  0.95 \\
horizon length $H$ & 96 & 96 & 96 & 96 \\
replay buffer size  & 1000000& 1000000 & 1000000 & 50000 \\
Optimizer & Adam & Adam & Adam & Adam  \\

\bottomrule
\end{tabular}

\end{table*}

.

\begin{table*}[ht!]
\scriptsize
\centering
\caption{Hyerparameters of ABPT no 0-step Value no Entropy}
\label{tab:comparision}
\begin{tabular}{l |  l | l | l | l}
\toprule
& \textbf{Hovering} & \textbf{Tracking} & \textbf{Landing} & \textbf{Racing} \\
\midrule
learning rate $\alpha$ & 0.01 & 0.01 & 0.002 & 0.002  \\
number of parallel environments $n$ & 100 & 100 & 100 & 100  \\
discount factor $\gamma$ & 0.99 & 0.99 & 0.99 & 0.99  \\
training critic steps per minibatch & 10 & 10 & 10 & 10 \\ 
weight decay  & 0.00001 & 0.00001 & 0.00001 & 0.00001 \\ 
target critic update factor $\tau$ & 0.005 & 0.005 & 0.005 & 0.005 \\ 
decay learning rate  & False & True & False & True \\ 
value estimation factor $\lambda$ &  0.95 &  0.95 &  0.95 &  0.95 \\
horizon length $H$ & 96 & 96 & 96 & 96 \\
replay buffer size  & 1000000& 1000000 & 1000000 & 50000 \\
Optimizer & Adam & Adam & Adam & Adam  \\

\bottomrule
\end{tabular}

\end{table*}

%% file: tables/r_robust.tex
\begin{table*}[ht!]
\scriptsize
\centering
\caption{Hyerparameters of ABPT upon Velocity-based Reward in Racing }
\label{tab:comparision}
\begin{tabular}{l |  l | l |  l | l}
\toprule
  & \textbf{ABPT}& \textbf{SHAC}& \textbf{BPTT}& \textbf{PPO} \\
\midrule
learning rate $\alpha$  & 0.02 & 0.02   & 0.002 & 0.0002 \\
number of parallel environments $n$ & 100 & 100 & 100 & 100  \\
discount factor $\gamma$  & 0.99 & 0.99 & 0.99 & 0.99  \\
training critic steps per minibatch  & 10  & 10 & 10 & 5\\ 
weight decay  & 0.00001  & 0.00001 & 0.00001 & 0.00001 \\ 
target critic update factor $\tau$  & 0.005& 0.005& - &- \\ 
decay learning rate  & True  & True  & True & -\\ 
value estimation factor $\lambda$  &  0.95   &  0.95 & -&-\\
horizon length $H$& 96 & 96 & 512 & -\\
replay buffer size  & 50000 &- &- &- \\
minibatch size & - & - & - & 51200  \\
GAE  &- &- &- & 1 \\
Optimizer & Adam & Adam & Adam & Adam  \\

\bottomrule
\end{tabular}

\end{table*}